\documentclass[11pt]{article}

\usepackage[preprint]{acl}
\usepackage{times}
\usepackage{latexsym}
\usepackage[T1]{fontenc}
\usepackage[utf8]{inputenc}
\usepackage{microtype}
\usepackage{inconsolata}
\usepackage{graphicx}
\usepackage{amsmath}
\usepackage{amssymb}
\usepackage{booktabs}
\usepackage{multirow}
\usepackage{makecell}
\usepackage{listings}

\usepackage{colortbl}

\definecolor{PerfBest}{HTML}{95d5b2}
\definecolor{PerfSecond}{HTML}{a3cef1}

\newcommand{\best}[1]{\cellcolor{PerfBest!55}\textbf{#1}}
\newcommand{\second}[1]{\cellcolor{PerfSecond!25}#1}
\newcommand{\modelgroup}[2]{\rowcolor{gray!20}\multicolumn{#1}{l}{\textbf{\textit{#2}}}}

\title{V-Rubrics: Visual Faithfulness via Rubric-Based Reinforcement Learning}

\author{%
  Shulin Tian$^{1,2*}$, Minglun Li$^{3*}$, Yuhao Dong$^{1\ddag*}$, Hao Ding$^{3*}$, \\
  \bf Jiarui Yao$^{4}$, Haiwen Diao$^{1}$, Jingkang Yang$^{1}$, Hongyuan Zhu$^{2}$, Ziwei Liu$^1$ \\ \\
  $^{1}$S-Lab, Nanyang Technological University, $^{2}$A*STAR, $^{3}$ Independent Researcher, $^{4}$UIUC \\
  $^*$Equal Contributions. $^{\ddag}$ Project Lead.\\
  \small\texttt{\{shulin002, yuhao013, ziwei.liu\}@ntu.edu.sg}
  \\
  \tt\normalsize\color{magenta}\url{https://shulin16.github.io/v-rubrics/}
}

\begin{document}
\maketitle

\begin{abstract}
Vision-language models can produce fluent answers that are \emph{insufficiently grounded in the visual evidence}: a single unsupported object, chart value, or intermediate inference can undermine an otherwise plausible response. We argue that this is a \emph{credit-assignment failure} in multimodal post-training. Scalar outcome rewards indicate whether an answer is acceptable, but do not identify which visual facts are grounded, which reasoning steps are valid, or which instruction constraints are missed. We introduce \textbf{Visual Rubrics-Based Reinforcement Learning}, which decomposes reference responses into atomic propositions and scores generated answers along Visual Faithfulness (VF), Reasoning Consistency (RC), and Instruction Following (IF). The resulting rubric items provide structured partial credit and localize rubric credit when supporting evidence spans are available. We first obtain an SFT checkpoint by fine-tuning Qwen3-VL-8B-Instruct on the public OpenMMReasoner-SFT-874K corpus, adapting OpenMMReasoner's cold-start data recipe. We construct \textbf{V-Rubrics 50K}, a 50,248-example training set from 17 visually grounded sources, by applying rule-based filters before deriving example difficulty from rejection-sampling scores and then annotating every example with Gemini-3-Pro under the same structured prompt and protocol. We train our model based on the same SFT checkpoint using component-wise, prefix-localized rubric credit. Experiments show that our rubric-based GRPO improves over both the shared SFT baseline and answer-only GRPO, with the largest gains on knowledge-oriented and visually grounded reasoning benchmarks. The results show rubrics as a useful reward abstraction for visual post-training.
\end{abstract}

\section{Introduction}
\label{sec:intro}

\begin{figure}[ht]
    \centering
    \includegraphics[width=\linewidth]{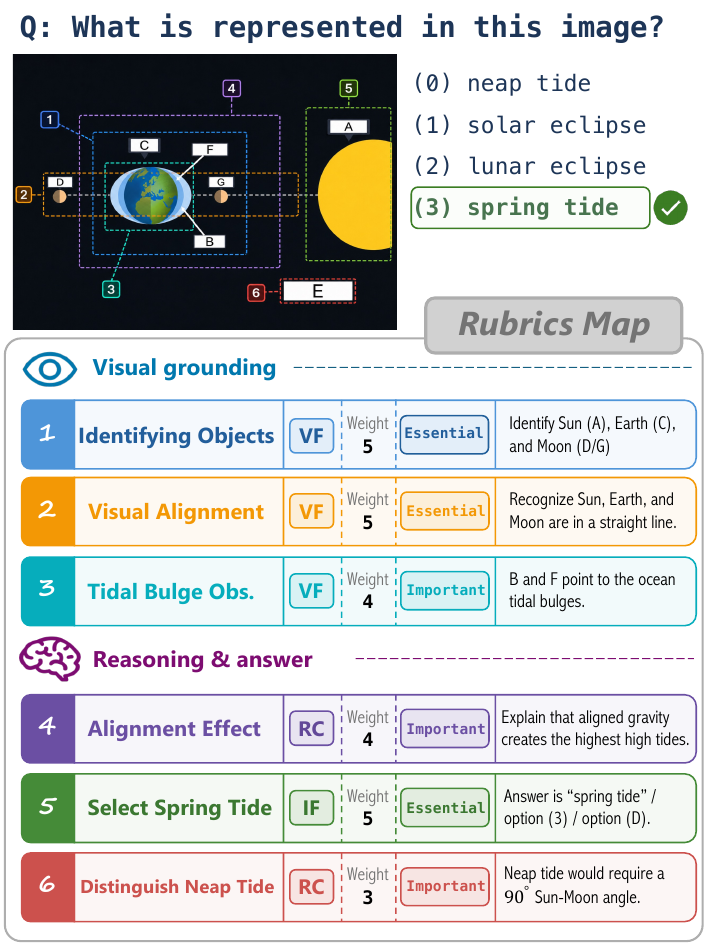}
    \caption{\textbf{Rubric-grounded visual reasoning.} A visual question answering example is decomposed into rubric items that check visual faithfulness, reasoning consistency, and instruction following.}
    \label{fig:example}
    \vspace{-16pt}
\end{figure}
\begin{figure*}[ht]
    \centering
    \includegraphics[width=\linewidth]{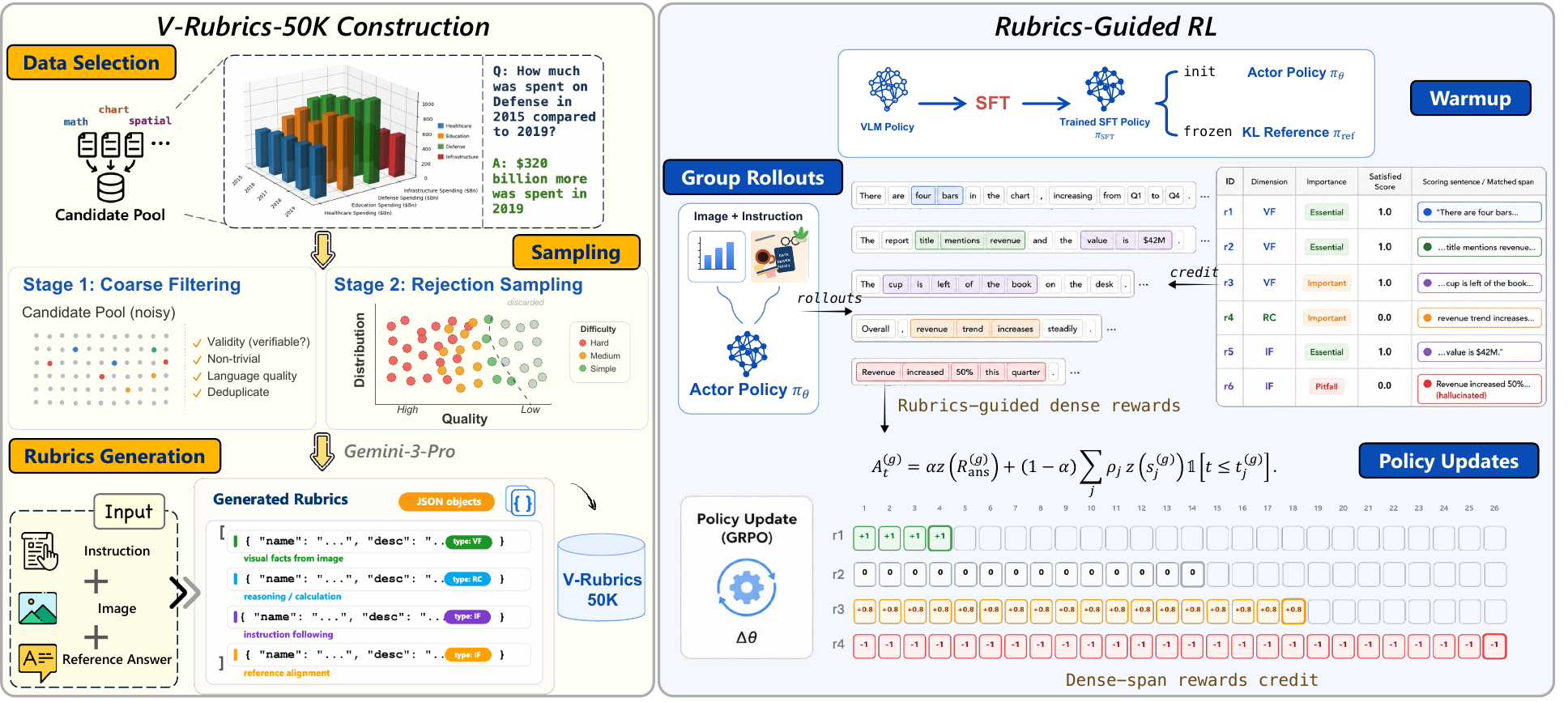}
    \caption{\textbf{Overview of Visual Rubrics-Based Reinforcement Learning.} V-Rubrics 50K expands VQA to atomic VF/RC/IF rubric items, which provide fine-grained scores and prefix-localized credit for GRPO-based VLM post-training.}
    \label{fig:teaser}
    \vspace{-10pt}
\end{figure*}

Vision-language models (VLMs) are increasingly used to answer questions, follow instructions, and generate explanations grounded in images. In these settings, fluent text is not enough: a useful response must be visually faithful, with objects, attributes, relations, counts, and inferred conclusions supported by image evidence. This requirement is especially demanding for charts, documents, diagrams, and dense scenes, where a single unsupported visual claim can change the final answer. Prior work on object hallucination~\citep{rohrbach-etal-2018-object,guan2024hallusionbench}, multimodal faithfulness evaluation~\citep{li-etal-2023-evaluating,jing-etal-2024-faithscore}, and benchmark-driven visual reasoning failures~\citep{wang2024charxiv} has shown that VLMs often produce plausible but unsupported details, and that conventional captioning or instruction-following metrics can miss these errors.

We view this failure not only as an evaluation problem, but as a credit-assignment problem for post-training. Standard RL alignment pipelines typically reduce a response to a scalar preference, a binary correctness label, or a holistic judge score. Such rewards are effective when correctness is directly verifiable~\citep{lambert2024tulu3,deepseek2025r1}, but visual reasoning often contains mixed evidence: a response may correctly identify relevant objects, make one unsupported inference, and still satisfy the requested format. A single outcome reward cannot say which visual claim is grounded, which reasoning step fails, or which instruction constraint is violated. Recent rubric-based RL shows that instance-specific criteria can extend reward learning beyond strictly verifiable tasks~\citep{gunjal2025rubrics}; for VLMs, the key question is how to make those criteria visually grounded and useful for optimization.

We propose \textbf{Visual Rubrics-Based Reinforcement Learning}, a framework that turns visually grounded, instance-specific criteria into fine-grained reinforcement-learning signals. For each image--instruction pair, we decompose the reference response into atomic rubric items along three dimensions: Visual Faithfulness (VF), Reasoning Consistency (RC), and Instruction Following (IF). VF checks whether stated content is supported by the image, RC checks whether conclusions follow from observed visual evidence, and IF checks whether the response satisfies the prompt requirements. Figure~\ref{fig:example} illustrates this decomposition on a single visual question. Rather than treating a response as an indivisible outcome, the framework provides partial credit for grounded content and targeted penalties for unsupported or inconsistent claims. Moreover, instead of collapsing rubric judgments into a single sequence-level score, our method preserves item-level reward components and, when supporting evidence can be aligned, localizes their advantages to the corresponding response prefixes, as summarized in Figure~\ref{fig:teaser}.

To support this framework, we construct \textbf{V-Rubrics 50K}, a training set of 50,248 examples from 17 visually grounded sources. Each example pairs an image, instruction, and reference response with atomic VF/RC/IF criteria and importance weights, turning visual grounding from a post-hoc diagnostic into structured supervision for post-training. Starting from a shared supervised checkpoint, we train the model using component-wise, prefix-localized rubric credit. Across general, knowledge-oriented, visual-mathematical, chart, and logical reasoning benchmarks, rubric-based GRPO improves over both the shared SFT baseline and answer-only GRPO, with the largest gains on tasks that depend on grounded intermediate reasoning. Ablations further favor the combined component-wise, prefix-localized design over scalar sequence-level rubric aggregation.

In summary, this work makes three contributions. First, we formulate visual faithfulness as a fine-grained RL credit-assignment problem and introduce a training framework that preserves item-level rubric components and, when aligned supporting evidence is available, localizes their advantages to response prefixes. Second, we introduce V-Rubrics 50K, a 50,248-example resource that turns reference responses into visually grounded atomic criteria with explicit capability dimensions and importance weights. Third, we show that rubric-based GRPO improves over both the shared SFT baseline and answer-only GRPO, with the strongest gains on tasks that require grounded intermediate reasoning; ablations further support the combined component-wise, prefix-localized credit design. More broadly, these results suggest that future VLM post-training should treat visual grounding not only as an evaluation diagnostic, but as a structured training interface that exposes where reasoning succeeds or fails.

\section{Related Work}
\label{sec:related}

\paragraph{Multimodal reasoning and reward granularity.}
Recent VLMs provide strong perception and instruction-following backbones for chart, document, diagram, and general visual reasoning~\citep{openai2024gpt4o,bai2025qwen25vl,bai2025qwen3vl}~\citep{zhu2025internvl3,li2024llavaonevision,an2025llavaonevision15}. The next question is how post-training should assign credit when a visual answer mixes correct observations, invalid inferences, and formatting constraints. Multimodal RLVR methods adapt verifiable or perception-oriented rewards to visual tasks~\citep{liu2025visualrft,shen2025vlmr1,huang2025visionr1}~\citep{xiao2025perceptionr1,ni2025pointrft}, while recent reasoning systems improve data mixtures, rollout selection, long reasoning traces, or on-policy optimization~\citep{chen2025vlaathinking,wang2025thinklite,zhang2025openmmreasoner}~\citep{leng2025mmr1,wei2025openvisionreasoner,feng2026onethinker}. These approaches establish that RL can improve multimodal reasoning, but the reward is usually still attached to the whole response or final answer. Our work studies a missing intermediate signal: a reward unit that identifies which visual facts are grounded, which reasoning steps are valid, and which instruction constraints are satisfied.

\paragraph{Rubrics as a reward interface.}
Rubric and judge-based evaluation is a natural way to define such intermediate units because it replaces a single correctness bit with explicit criteria. Text-side judges show that criteria can make automatic evaluation more interpretable and better aligned with human preferences~\citep{liu-etal-2023-g,kim-etal-2024-prometheus2,hashemi-etal-2024-llm}; multimodal judges and reward models extend the same idea to image-conditioned answers~\citep{ge2023mllmbench,lee-etal-2024-prometheus-vision,xiong2025llavacritic}~\citep{wang2025skyworkvlreward,li2024vlrewardbench,yasunaga2025multimodalrewardbench}. Rubrics as Rewards is the closest training analogue, showing that instance-specific rubrics can support on-policy RL beyond strictly verifiable domains~\citep{gunjal2025rubrics}. V-Rubrics makes this interface visually grounded: VF and RC criteria are tied to image evidence or licensed inference, while IF criteria capture prompt constraints. Satisfaction scores and item weights determine how each criterion contributes to reward and credit assignment.

\paragraph{From alignment feedback to local credit.}
Our training loop also relates to multimodal alignment methods that use human or AI feedback to improve reliability. LLaVA-RLHF, RLHF-V, fine-grained AI feedback, HDPO, and RLAIF-V show that preference, critique, or judge signals can shape VLM behavior after supervised tuning~\citep{sun-etal-2024-aligning,yu2024rlhfv,xiao2024fgaif}~\citep{fu-etal-2025-hdpo,yu2025rlaifv}. V-Rubrics differs in the form of feedback it exposes to the optimizer: instead of a holistic preference or critique, the training signal is decomposed into visually grounded propositions for faithfulness, reasoning consistency, and instruction following. This decomposition enables prefix-localized, item-factorized credit. Broader hallucination diagnostics, long-chain visual reasoning, video reasoning, and in-context adaptation work are discussed in Appendix~\ref{app:more_related_work}.

\section{Method}
\label{sec:method}

We propose Visual Rubrics-Based Reinforcement Learning, a training framework that converts fine-grained visual rubrics into item-level scores and prefix-localized token advantages. The complete training pipeline is organized into five parts: the problem setup, SFT initialization, V-Rubrics 50K construction, rubric design, and rubric-based RL training.

\subsection{Problem Setup}

Let $x=(v,q)$ denote an image-instruction pair, where $v$ is an image and $q$ is a user instruction or question. A VLM policy $\pi_\theta$ generates a response $a \sim \pi_\theta(\cdot \mid x)$. A reference response $y$ provides the target content, but we do not treat $y$ as a single indivisible answer. For each $x$, let $\mathcal{I}(x)=\{1,\ldots,m(x)\}$ denote the rubric-index set. We decompose $y$ into
\begin{equation}
\mathcal{R}(x)=\{(r_j,d_j,c_j,w_j)\}_{j\in\mathcal{I}(x)},
\end{equation}
where $r_j$ is a self-contained atomic criterion (including its item name and description), $d_j \in \{\mathrm{VF}, \mathrm{RC}, \mathrm{IF}\}$ is its rubric dimension, $c_j$ is an importance label, and $w_j$ is the corresponding numeric item weight. The dimensions correspond to Visual Faithfulness, Reasoning Consistency, and Instruction Following. Each training record pairs $x$ with a fixed reference $y$; the later reward notation conditions on this associated reference implicitly.

The goal is to optimize $\pi_\theta$ so that generated responses are faithful to the image, logically consistent, and aligned with the user instruction. Our main training objective retains these item-level judgments during advantage construction rather than immediately collapsing them into a single outcome score.
This formulation is useful because visual answers often contain multiple claims with different grounding status. A response can mention the right objects but infer the wrong relation, solve the reasoning step but ignore the requested format, or reach the correct final answer through an unsupported visual shortcut. A single outcome score against the reference cannot distinguish these cases, whereas rubric items define the units on which credit should be assigned.

\subsection{SFT Initialization and Pipeline}
\label{sec:sft_initialization}

We begin with Qwen3-VL-8B-Instruct~\citep{qwen3vl8b,bai2025qwen3vl} and adapt the cold-start data recipe of OpenMMReasoner~\citep{zhang2025openmmreasoner} to this backbone. We fine-tune the backbone on the public OpenMMReasoner-SFT-874K corpus. We denote the resulting policy by $\pi_{\mathrm{SFT}}$. It is fixed before V-Rubrics construction and generates the rollouts used for rejection sampling. For each subsequent GRPO run, we create two independent parameter copies: $\pi_{\theta^{(0)}}$ initializes the trainable actor, whereas $\pi_{\mathrm{ref}}$ is the frozen KL reference. As policies at initialization,
\begin{equation}
\pi_{\theta^{(0)}}=\pi_{\mathrm{ref}}=\pi_{\mathrm{SFT}},
\qquad \pi_{\mathrm{ref}}\ \text{remains frozen}.
\label{eq:policy_initialization}
\end{equation}
Thus, the same supervised policy underlies data selection and both RL comparisons, while the actor and reference remain distinct parameter copies during RL.

\subsection{V-Rubrics 50K Construction}

V-Rubrics 50K is built from visually grounded training datasets covering diagram reasoning~\citep{kembhavi2016ai2d,lu2021intergps}, chart understanding~\citep{li2025chartgalaxy,masry2022chartqa,hegde2025chartqax,xia2024chartx}, document VQA~\citep{mathew2021docvqa,mathew2021infographicvqa}, mathematical visual reasoning~\citep{qiao2025wemath2}, counting~\citep{deitke2025molmo}, educational QA~\citep{du2025mmprm}, and general visual reasoning~\citep{wang2025thinklite,wang2025vlrethinker}. We first normalize examples from these source datasets into a common image--instruction--reference format. We then apply deterministic rule-based filters to retain records with valid media and required fields, non-trivial task content, and adequate language quality, while removing identity- and strict-content duplicates. For each filtered example, the fine-tuned $\pi_{\mathrm{SFT}}$ generates eight rejection-sampling rollouts. If $k$ of the eight rollouts are judged correct, the empirical success rate $k/8$ provides a model-relative signal of example difficulty. We map this signal to hard, medium, and simple categories and use these assignments to form a fixed training set of 50,248 examples from 17 canonical sources. The final mixture contains 18,121 hard (36.1\%), 25,306 medium (50.4\%), and 6,821 simple examples (13.6\%).

\begin{figure}[htbp]
    \centering
    \includegraphics[width=\linewidth]{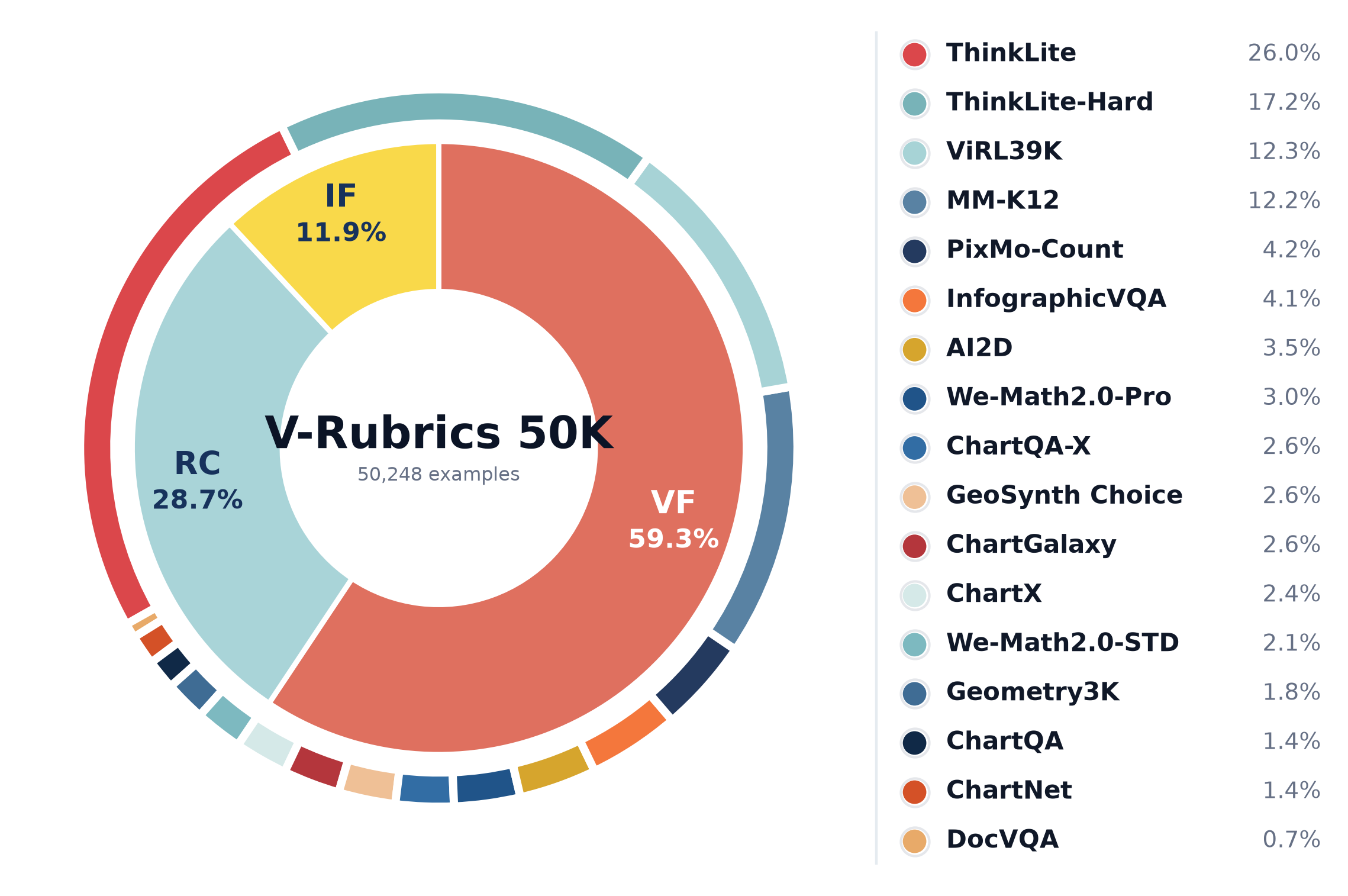}
    \caption{\textbf{Data Distribution of V-Rubrics 50K.} The inner ring shows the VF/RC/IF composition of 352,938 rubric items, while the thin outer ring shows the distribution of 50,248 examples across 17 canonical sources. Outer-ring arc length is proportional to each source's percentage, with colors mapped to source names in the legend.}
    \label{fig:data_distribution}
    \vspace{-10pt}
\end{figure}

Finally, we turn each selected example into rubric-based supervision. Every example is annotated by \texttt{Gemini-3-Pro} using the same structured multimodal prompt and annotation protocol~\citep{google2026gemini3pro}. The protocol converts the image, instruction, and reference response into atomic VF/RC/IF criteria with explicit importance weights and a shared output schema. Figure~\ref{fig:data_distribution} summarizes the resulting data distribution, and Appendix~\ref{app:data_construction} and Table~\ref{tab:vrubrics_source_counts} report the construction details and final source inventory.

\subsection{Rubric Design Principles}
\label{sec:rubric_design}

Following recent work that treats rubrics as reusable reward functions for domains without simple correctness checks \citep{gunjal2025rubrics}, we impose four design principles on V-Rubrics 50K. Rubrics must be \textbf{visually grounded}, so VF items refer to checkable image evidence such as objects, attributes, relations, counts, visible text, or chart values. They must be \textbf{self-contained}, so each criterion states the target explicitly and the verifier can score the response from the criterion text without reconstructing the intended standard from the full reference answer. They must provide \textbf{coverage}, so the item set evaluates intermediate visual facts and reasoning steps rather than only the final answer. Finally, they must encode \textbf{importance}, so central criteria receive greater weight; example difficulty is represented separately by a category derived from the stored rejection-sampling score and used to describe and balance the data mixture.

We operationalize these principles through a structured rubric-generation prompt that produces atomic JSON criteria with an importance prefix, numeric weight, and dimension label. Each item is labeled \textsc{Essential}, \textsc{Important}, \textsc{Optional}, or \textsc{Pitfall}: the first three labels reward required or useful grounded content, while pitfall labels identify common hallucinations or reasoning traps. Appendix~\ref{app:rubric_prompt} gives the rubric-generation instructions and output schema, and Appendix~\ref{app:detailed_reward_training} describes how these labels are mapped into training rewards.
Although the rubrics are generated automatically, the schema is designed to make verification easier than free-form grading. Each criterion is short, self-contained, and tied to a single checkable proposition, so the verifier does not need to infer the intended evaluation standard from the whole answer. This reduces dependence on a holistic judge preference and makes the reward auditable at the item level.

\subsection{Reward Design and RL Training}
\subsubsection{Rubric-Based Reward}

Given a response $a$, an LLM rubric verifier independently assigns each criterion an aligned binary satisfaction score $s_j(a;x)\in\{0,1\}$. Here $x$ selects the example-specific criterion $r_j$; the verifier receives $a$ and $r_j$, not the raw image. Larger values always indicate better compliance. In particular, a \textsc{Pitfall} criterion is written as a desired avoidance condition, so $s_j=1$ means that the response avoids the described failure.

Let $\mathcal{I}_{+}(x)=\{j\in\mathcal{I}(x):w_j>0\}$ denote the positive criteria. Each example contains at least one such criterion. Positive criteria provide importance-weighted partial credit:
\begin{equation}
\begin{aligned}
R_{\mathrm{rub}}(a,x)
&=\sum_{j\in\mathcal{I}_{+}(x)} \rho_j\,s_j(a;x),\\
\rho_j
&=\frac{w_j}{\sum_{k\in\mathcal{I}_{+}(x)}w_k},
\quad j\in\mathcal{I}_{+}(x).
\end{aligned}
\label{eq:sequence_rubric_reward}
\end{equation}
The weights are normalized once over the positive criteria. A confirmed \textsc{Pitfall} violation acts as a semantic veto on answer and positive-rubric credit; we keep this gate implicit below.

At the semantic level, we blend the rubric signal with a final-answer reward $R_{\mathrm{ans}}(a,x)\in\{0,1\}$ using $\alpha\in[0,1]$:
\begin{equation}
R(a,x)=\alpha\,R_{\mathrm{ans}}(a,x)+(1-\alpha)\,R_{\mathrm{rub}}(a,x).
\label{eq:blended_reward}
\end{equation}
Training also uses a standard binary format reward, omitted here for clarity. The answer term anchors task success, while rubric items identify which parts of a response support it. Appendix~\ref{app:detailed_reward_training} details the verifier and the VF/RC/IF categories.

\subsubsection{Rubrics-Guided RL Training}
Both variants use \textbf{Group Relative Policy Optimization (GRPO)} \citep{shao2024deepseekmath} and differ in how feedback is converted into advantages. For each $x$, GRPO compares $G$ rollouts $a^{(g)}=(a_1^{(g)},\ldots,a_{T_g}^{(g)})$ sampled from $\pi_{\theta_{\mathrm{old}}}$. For any score component $u$, we define its group-relative standardization as
\begin{equation}
z\big(u^{(g)}\big)=\frac{u^{(g)}-\mu_u}{\sigma_u+\epsilon_z},
\label{eq:group_baseline}
\end{equation}
where $\mu_u$ and $\sigma_u$ are computed over rollouts with an available value for $u$, and $\epsilon_z$ is a numerical-stability constant. Appendix~\ref{app:grpo_optimization} gives the complete clipped objective and implementation details; below we focus on the construction of $A_t^{(g)}$.

\paragraph{From sequence-level to component-wise prefix credit.}
\label{sec:dense_credit}
Let $R^{(g)}:=R(a^{(g)},x)$ and $R_{\mathrm{ans}}^{(g)}:=R_{\mathrm{ans}}(a^{(g)},x)$. The \emph{sequence-level} variant assigns $z(R^{(g)})$ to the entire response, recovering standard outcome-supervised GRPO and serving as one of our ablations.

Rubric feedback carries more structure: the verifier scores each item separately and, when available, returns the response sentence supporting its decision. Write $s_j^{(g)}:=s_j(a^{(g)};x)$, and let $t_{j,\mathrm{end}}^{(g)}$ be the final token in the aligned evidence span. If no reliable span is available for a scored item, we set $t_{j,\mathrm{end}}^{(g)}=T_g$. The resulting prefix mask is
\begin{equation}
M_{j,t}^{(g)}
=
\mathbf{1}\!\left[t\le t_{j,\mathrm{end}}^{(g)}\right].
\label{eq:prefix_mask}
\end{equation}
Let $\mathcal{I}_{\mathrm{sc}}^{(g)}=\{j\in\mathcal{I}_{+}(x):s_j^{(g)}\text{ is available}\}$ be the successfully scored positive rubric indices for rollout $g$. The answer advantage is broadcast across the full response, while each available rubric advantage contributes only through its prefix mask:
\begin{equation}
\begin{aligned}
A_t^{(g)}=\;
&\alpha\,z\big(R_{\mathrm{ans}}^{(g)}\big)\\
&+(1-\alpha)\sum_{j\in\mathcal{I}_{\mathrm{sc}}^{(g)}}\rho_j\,
z\big(s_j^{(g)}\big)\,
M_{j,t}^{(g)},
\end{aligned}
\label{eq:dense_credit}
\end{equation}
The answer component spans the full response, while each positive rubric is standardized separately and contributes only through its prefix. Item weights are normalized once and are not renormalized after masking. An unlocalized item receives sequence-wide, item-factorized credit; when all scored rollouts agree on an item, it contributes no gradient. Appendix~\ref{app:detailed_reward_training} gives the localization and missing-judgment details.

\begin{table*}[!htbp]
\centering
\scriptsize
\renewcommand{\arraystretch}{0.66}
\setlength{\tabcolsep}{3.8pt}
\caption{\textbf{Performance on general VLM and knowledge benchmarks.}
Within each model group, green and blue mark the best and second-best results;
Knowledge Avg. is the mean of the three knowledge metrics, while Overall Avg.
additionally includes MMBench-Dev.}
\label{tab:rubrics_main_general_knowledge}

\resizebox{\textwidth}{!}{%
\begin{tabular}{lcc c cccc c}
\toprule
\multirow{2}{*}{\textbf{Model}}
& \multirow{2}{*}{\textbf{SFT Data}}
& \multirow{2}{*}{\textbf{RL Data}}
& \multicolumn{1}{c}{\textbf{General VLM}}
& \multicolumn{4}{c}{\textbf{Knowledge \& Academic Reasoning}}
& \multirow{2}{*}{\textbf{Overall Avg.}} \\
\cmidrule(lr){4-4}\cmidrule(lr){5-8}
 &  &  &
\textbf{MMBench-Dev}
 & \textbf{MMMU Val}
 & \textbf{MMMU-Pro}
 & \textbf{MMMU-Pro V}
 & \textbf{Avg.}
 & \\
\midrule

\modelgroup{9}{Closed-source models} \\
\midrule
GPT-4o~\citep{openai2024gpt4o} & -- & -- & \best{88.4} & \best{69.1} & \best{54.0} & \best{49.7} & \best{57.60} & \best{65.30} \\
GPT-4o mini~\citep{openai2024gpt4omini} & -- & -- & \second{83.8} & \second{59.4} & \second{39.9} & \second{35.2} & \second{44.83} & \second{54.58} \\

\midrule
\modelgroup{9}{Open-source Instruct models} \\
\midrule
LLaVA-OneVision-7B~\citep{li2024llavaonevision} & 4.8M & -- & 80.8 & 48.8 & 29.5 & 18.7 & 32.33 & 44.45 \\
InternVL3-8B~\citep{zhu2025internvl3} & -- & -- & 83.6 & \second{62.7} & -- & -- & -- & -- \\
Qwen2.5-VL-7B~\citep{bai2025qwen25vl} & -- & -- & \best{87.8} & 58.6 & 37.9 & 35.1 & \second{43.87} & \second{54.85} \\
Qwen3-VL-8B-Instruct$^\dagger$~\citep{bai2025qwen3vl} & -- & -- & \second{86.08} & \best{69.00} & \best{57.75} & \best{57.69} & \best{61.48} & \best{67.63} \\
LLaVA-OneVision-1.5-8B~\citep{an2025llavaonevision15} & 105M & -- & 84.14 & 55.4 & 37.4 & 25.2 & 39.33 & 50.54 \\
OMR-7B-ColdStart~\citep{zhang2025openmmreasoner} & 874k & -- & -- & 54.4 & \second{39.3} & \second{37.3} & -- & -- \\

\midrule
\modelgroup{9}{Open-source Thinking models} \\
\midrule
VLAA-Thinker-Qwen2.5-7B~\citep{chen2025vlaathinking} & 126k & 25k & -- & -- & -- & -- & -- & -- \\
ThinkLite-7B-VL~\citep{wang2025thinklite} & -- & 11k & 81.4 & -- & -- & -- & -- & -- \\
VL-Rethinker-7B~\citep{wang2025vlrethinker} & -- & 39k & -- & -- & 41.7 & -- & -- & -- \\
M2-Reasoning~\citep{wang2025m2reasoning} & 6.2M & 102k & -- & -- & -- & -- & -- & -- \\
MMR1~\citep{leng2025mmr1} & 1.6M & 15k & \second{86.9} & 52.4 & 41.1 & 37.1 & 43.53 & 54.38 \\
OpenVLThinker-7B~\citep{deng2025openvlthinker} & 3.3k & 9.6k & 81.3 & 55.1 & 39.7 & 38.4 & 44.40 & 53.63 \\
MM-Eureka-Qwen-7B~\citep{meng2025mmeureka} & -- & 15.6k & 79.3 & 54.4 & 40.1 & 37.1 & 43.87 & 52.73 \\
OVR-7B~\citep{wei2025openvisionreasoner} & 2M & 300k & -- & 51.8 & \second{50.2} & 29.1 & -- & -- \\
OMR-7B~\citep{zhang2025openmmreasoner} & 874k & 74k & 85.9 & 57.8 & 44.1 & \second{40.6} & \second{47.50} & \second{57.10} \\
OneThinker-8B~\citep{feng2026onethinker} & 340k & 600k & 86.6 & \second{70.6} & -- & -- & -- & -- \\
Qwen3-VL-8B-Thinking$^\dagger$~\citep{bai2025qwen3vl} & -- & -- & \best{87.29} & \best{72.22} & \best{59.48} & \best{59.19} & \best{63.63} & \best{69.55} \\

\midrule
\modelgroup{9}{Ours} \\
\midrule
SFT
 & 874k & -- & 84.79 & 66.78 & 54.34 & 53.82 & 58.31 & 64.93 \\
+ GRPO
 & 874k & 50k & \best{86.94} & \second{68.00} & \second{55.72} & \second{54.34} & \second{59.35} & \second{66.25} \\
+ GRPO w/ rubrics (Ours)
 & 874k & 50k & \second{86.51} & \best{70.56} & \best{58.15} & \best{56.94} & \best{61.88} & \best{68.04} \\
\bottomrule
\end{tabular}%
}
\vspace{-0.5em}
\end{table*}

\begin{table*}[ht]
\centering
\scriptsize
\renewcommand{\arraystretch}{0.92}
\setlength{\tabcolsep}{1.6pt}
\caption{\textbf{Performance on visual math, chart, and logic benchmarks.}
Within each model group, green and blue mark the best and second-best results;
Math Avg. is the mean of the five visual-math metrics, Chart Avg. the mean of
LogicVista and CharXiv, and Overall Avg. the mean of all seven metrics.}
\label{tab:rubrics_main_visual_math}

\resizebox{\textwidth}{!}{%
\begin{tabular}{lcc cccccc ccc c}
\toprule
\multirow{2}{*}{\textbf{Model}}
& \multirow{2}{*}{\textbf{SFT Data}}
& \multirow{2}{*}{\textbf{RL Data}}
& \multicolumn{6}{c}{\textbf{Visual Math \& Reasoning}}
& \multicolumn{3}{c}{\textbf{Chart \& Logic}}
& \multirow{2}{*}{\textbf{Overall Avg.}} \\
\cmidrule(lr){4-9}\cmidrule(lr){10-12}
 &  &  &
\makecell{\textbf{MathVista}\\\textbf{mini}}
 & \makecell{\textbf{MathVision}\\\textbf{test}}
 & \makecell{\textbf{MathVerse}\\\textbf{V/O}}
 & \makecell{\textbf{DynaMath}\\\textbf{Worst}}
 & \makecell{\textbf{WeMath}\\\textbf{Loose}}
 & \makecell{\textbf{Math}\\\textbf{Avg.}}
 & \makecell{\textbf{Logic}\\\textbf{Vista}}
 & \makecell{\textbf{CharXiv}\\\textbf{Reas.}}
 & \makecell{\textbf{Chart}\\\textbf{Avg.}}
 & \\
\midrule

\modelgroup{13}{Closed-source models} \\
\midrule
GPT-4o~\citep{openai2024gpt4o} & -- & -- & \best{63.8} & \best{31.1} & \best{40.6} & \best{34.5} & \best{62.8} & \best{46.56} & \best{64.4} & -- & -- & -- \\
GPT-4o mini~\citep{openai2024gpt4omini} & -- & -- & \second{55.1} & \second{27.3} & \second{30.0} & \second{31.6} & \second{48.8} & \second{38.56} & \second{41.4} & \best{34.1} & \best{37.75} & \best{38.33} \\

\midrule
\modelgroup{13}{Open-source Instruct models} \\
\midrule
LLaVA-OneVision-7B~\citep{li2024llavaonevision} & 4.8M & -- & 62.6 & 17.6 & 17.6 & 9.0 & 43.5 & 30.06 & 32.0 & 23.6 & 27.80 & 29.41 \\
InternVL3-8B~\citep{zhu2025internvl3} & -- & -- & 70.5 & 28.6 & 33.9 & 23.0 & 58.8 & 42.96 & 43.6 & 37.6 & 40.60 & 42.29 \\
Qwen2.5-VL-7B~\citep{bai2025qwen25vl} & -- & -- & 69.2 & 25.5 & 41.1 & 21.8 & 53.1 & 42.14 & \second{47.9} & 36.4 & 42.15 & 42.14 \\
Qwen3-VL-8B-Instruct$^\dagger$~\citep{bai2025qwen3vl} & -- & -- & \best{76.60} & \best{56.41} & \second{47.97} & \best{40.72} & \best{75.81} & \best{59.50} & \best{62.19} & \best{50.20} & \best{56.20} & \best{58.56} \\
LLaVA-OneVision-1.5-8B~\citep{an2025llavaonevision15} & 105M & -- & 69.6 & 25.6 & 46.3 & 19.8 & 49.4 & 42.14 & 45.8 & 37.0 & 41.40 & 41.93 \\
OMR-7B-ColdStart~\citep{zhang2025openmmreasoner} & 874k & -- & \second{74.8} & \second{36.6} & \best{57.7} & \second{29.3} & \second{67.2} & \second{53.12} & 46.2 & \second{39.7} & \second{42.95} & \second{50.21} \\

\midrule
\modelgroup{13}{Open-source Thinking models} \\
\midrule
VLAA-Thinker-Qwen2.5-7B~\citep{chen2025vlaathinking} & 126k & 25k & 68.0 & 26.4 & 48.2 & 22.4 & 61.7 & 45.34 & 48.5 & -- & -- & -- \\
ThinkLite-7B-VL~\citep{wang2025thinklite} & -- & 11k & 71.6 & 24.6 & 42.9 & 16.5 & -- & -- & 42.7 & -- & -- & -- \\
VL-Rethinker-7B~\citep{wang2025vlrethinker} & -- & 39k & \best{80.3} & 28.4 & 46.4 & 17.8 & -- & -- & 42.7 & -- & -- & -- \\
M2-Reasoning~\citep{wang2025m2reasoning} & 6.2M & 102k & 75.0 & 42.1 & 40.4 & -- & -- & -- & 50.6 & -- & -- & -- \\
MMR1~\citep{leng2025mmr1} & 1.6M & 15k & 72.0 & 31.8 & 55.4 & 27.9 & 68.0 & 51.02 & 48.9 & 43.5 & 46.20 & 49.64 \\
OpenVLThinker-7B~\citep{deng2025openvlthinker} & 3.3k & 9.6k & 65.3 & 23.0 & 38.1 & 16.8 & 61.9 & 41.02 & 44.5 & 41.0 & 42.75 & 41.51 \\
MM-Eureka-Qwen-7B~\citep{meng2025mmeureka} & -- & 15.6k & 72.6 & 28.1 & 45.4 & 23.0 & 59.8 & 45.78 & 46.3 & 42.4 & 44.35 & 45.37 \\
OVR-7B~\citep{wei2025openvisionreasoner} & 2M & 300k & 72.1 & \second{51.8} & 54.6 & 33.5 & 64.8 & 55.36 & \second{54.8} & 44.5 & \second{49.65} & 53.73 \\
OMR-7B~\citep{zhang2025openmmreasoner} & 874k & 74k & \second{79.5} & 43.6 & \second{63.8} & \second{34.9} & \second{79.0} & \second{60.16} & 50.0 & \second{46.1} & 48.05 & \second{56.70} \\
OneThinker-8B~\citep{feng2026onethinker} & 340k & 600k & -- & -- & \best{64.3} & -- & -- & -- & -- & -- & -- & -- \\
Qwen3-VL-8B-Thinking$^\dagger$~\citep{bai2025qwen3vl} & -- & -- & 77.80 & \best{62.70} & 52.03 & \best{40.32} & \best{84.67} & \best{63.50} & \best{63.53} & \best{54.50} & \best{59.02} & \best{62.22} \\

\midrule
\modelgroup{13}{Ours} \\
\midrule
SFT
 & 874k & -- & 78.30 & 55.46 & 47.84 & \second{41.32} & 77.43 & 60.07 & \second{60.63} & 48.20 & 54.42 & 58.45 \\
+ GRPO
 & 874k & 50k & \second{81.10} & \second{56.71} & \best{52.16} & 41.12 & \second{84.86} & \second{63.19} & \second{60.63} & \best{57.00} & \second{58.81} & \second{61.94} \\
+ GRPO w/ rubrics (Ours)
 & 874k & 50k & \best{81.30} & \best{58.88} & \second{49.37} & \best{42.32} & \best{86.29} & \best{63.63} & \best{62.42} & \second{56.60} & \best{59.51} & \best{62.45} \\
\bottomrule
\end{tabular}%
}
\vspace{-0.5em}
\end{table*}

\section{Experiments}
\label{sec:experiments}
We evaluate whether rubric-based rewards improve visually grounded reasoning while preserving the general capabilities of the underlying VLM. The results are reported in Tables~\ref{tab:rubrics_main_general_knowledge} and~\ref{tab:rubrics_main_visual_math}.

\subsection{Experimental Setup}

\paragraph{Model.}
All our models derive from Qwen3-VL-8B-Instruct~\citep{qwen3vl8b,bai2025qwen3vl}. For each RL run, the trainable actor is initialized as $\pi_{\theta^{(0)}}=\pi_{\mathrm{SFT}}$, and the separately instantiated $\pi_{\mathrm{ref}}=\pi_{\mathrm{SFT}}$ remains frozen as the KL reference policy, following Section~\ref{sec:sft_initialization}. We train with GRPO using \texttt{verl}~\citep{sheng2024verl} and apply the KL penalty in the actor loss. Appendix~\ref{app:hyperparams} gives the detailed training configuration.

\paragraph{Training.}
\textit{1) SFT data.} OpenMMReasoner-SFT-874K is the 874K-example cold-start mixture released with OpenMMReasoner~\citep{zhang2025openmmreasoner}. It combines five components---LLaVA-CoT, MiroMind-M1, filtered MMR1, OpenVLThinker-SFT-iter3, and WeMath---and is used to produce the shared SFT initialization described above.
\textit{2) RL data.} RL training uses the constructed V-Rubrics 50K, a 50,248-example collection of image-instruction-reference triples annotated with VF/RC/IF rubric items. Every example includes atomic propositions and signed importance weights together with an example-level \texttt{rs\_score} from which difficulty is derived. The fixed difficulty-stratified mixture contains 18,121 hard, 25,306 medium, and 6,821 simple examples. The answer-level and rubric-based GRPO variants use the same RL examples; their reward and credit-assignment mechanisms differ, while Appendix~\ref{app:hyperparams} reports the training configuration and the batch sizes used by each variant.
\textit{3) LLM-as-a-judge.} Training rewards are assigned by Qwen3-VL-235B-A22B, which serves as the LLM judge. We use two judge prompts: an answer-equivalence judge that compares the parsed answer with the reference answer, and a rubric verifier that independently scores the generated response against each self-contained VF/RC/IF criterion. For the prefix-credit run, the rubric verifier additionally returns the response sentence supporting its decision; when that sentence can be reliably aligned, it is used for token-level credit localization.

\paragraph{Evaluation.}
We evaluate with VLMEvalKit \citep{duan2024vlmevalkit} on 10 benchmark families covering general VLM ability, knowledge-oriented reasoning, visual math, chart reasoning, and logic. The suite includes MMBench~\citep{liu2024mmbench}, MMMU and MMMU-Pro~\citep{yue2024mmmu,yue2024mmmupro}, MathVista~\citep{lu2024mathvista}, MathVision~\citep{wang2024mathvision}, MathVerse~\citep{zhang2024mathverse}, DynaMath~\citep{zou2024dynamath}, WeMath~\citep{qiao2024wemath}, LogicVista~\citep{xiao2024logicvista}, and CharXiv~\citep{wang2024charxiv}. We report standard accuracy and unweighted averages over the displayed metrics and splits.

\subsection{Baselines}

Tables~\ref{tab:rubrics_main_general_knowledge} and~\ref{tab:rubrics_main_visual_math} compare our models against closed-source, open-source instruct, and open-source thinking baselines, including Qwen3-VL-8B-Instruct and Qwen3-VL-8B-Thinking evaluated by us. For our models, \textbf{SFT} is our Qwen3-VL-8B checkpoint trained on OpenMMReasoner-SFT-874K, \textbf{+ GRPO} adds scalar answer-level RL to that checkpoint, and \textbf{+ GRPO w/ rubrics (Ours)} augments the answer-level signal with component-wise, prefix-localized V-Rubrics 50K credit using the shared difficulty-stratified mixture in Section~\ref{sec:method}. Both RL variants share the supervised initialization, RL dataset, rollout budget, and optimizer settings, differing only in their batch sizes and in the reward and credit assignment (Appendix~\ref{app:hyperparams}).

\subsection{Main Results}

Tables~\ref{tab:rubrics_main_general_knowledge} and~\ref{tab:rubrics_main_visual_math} show two consistent trends. First, online RL improves the Qwen3-VL-8B training stack, but rubric rewards make the improvement more targeted. On the general/knowledge table, answer-level GRPO gives a modest gain over SFT, while augmenting the answer-level reward with rubric credit yields an additional $1.79$-point improvement in Overall Avg. (about $2.7\%$ relative). The gains concentrate on MMMU and MMMU-Pro rather than MMBench-Dev, where scalar GRPO is slightly higher. This suggests that rubrics mainly help when the benchmark rewards multi-step academic reasoning rather than broad VLM capability alone.

Second, the effect is clearer on visual math, chart, and logic tasks. Rubric-based GRPO improves the visual-reasoning Overall Avg. by $4.00$ points over SFT (about $6.8\%$ relative) and remains slightly ahead of answer-level GRPO. The strongest gains appear on metrics that depend on grounded intermediate perception, such as MathVision, DynaMath, WeMath, and LogicVista. At the same time, scalar GRPO remains better on MathVerse V/O and CharXiv reasoning, so the improvement is not a uniform benchmark lift; it is concentrated where rubric items can expose useful partial credit.

Overall, the main result is that structured rubric rewards improve over the SFT and scalar-GRPO baselines most reliably on reasoning-heavy visual tasks, while largely preserving general VLM performance. Section~\ref{sec:analysis} analyzes why this pattern emerges from the reward design.

\paragraph{Ablations.}
Table~\ref{tab:ablations} compares the answer-only baseline with two answer-plus-rubric variants: scalar sequence-level aggregation and component-wise, prefix-localized advantage composition.
\begin{table}[t]
\vspace{-0.4em}
\centering
\footnotesize
\setlength{\tabcolsep}{3pt}
\renewcommand{\arraystretch}{1.08}
\begin{tabular}{@{}llc@{}}
\toprule
\textbf{Signal} & \textbf{Credit} & \textbf{Overall Avg.} \\
\midrule
Answer only      & Sequence-level     & 66.25 \\
Answer + rubrics & Sequence-level     & 67.74 \\
Answer + rubrics & Component + prefix & \best{68.04} \\
\bottomrule
\end{tabular}
\caption{\textbf{Reward and credit-assignment ablation.}}
\label{tab:ablations}
\vspace{-0.6em}
\end{table}

Relative to the answer-only baseline at $66.25$, scalar sequence-level rubric aggregation improves Overall Avg. by $1.49$ points, and component-wise prefix credit improves it by $1.79$ points. The component-wise prefix variant reaches $68.04$ versus $67.74$ for scalar rubric aggregation; because it changes both component-wise standardization and localization, this additional $0.30$-point difference reflects their combined effect rather than localization alone. We use this variant for our main rubric-based model.

\subsection{Analysis}
\label{sec:analysis}
\begin{figure}[t]
\centering
\includegraphics[width=\columnwidth]{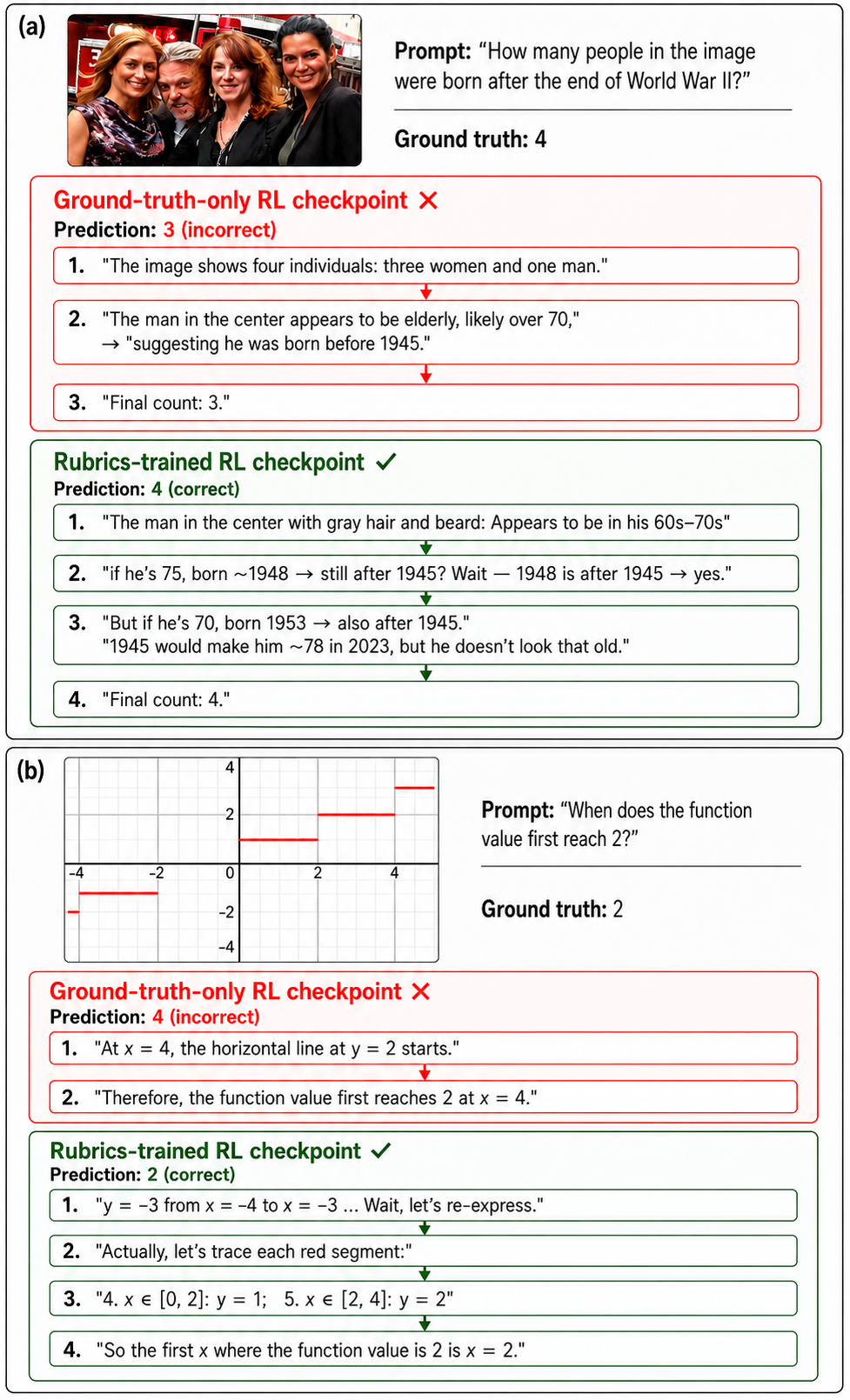}
\caption{\textbf{Qualitative comparison between answer-only GRPO and rubric-based checkpoints.} Top: rubric training
improves reasoning consistency by making the intermediate inference explicit. Bottom: rubric training improves visual faithfulness by tracing the graph before answering.}
\label{fig:qualitative}
\vspace{-10pt}
\end{figure}

\paragraph{Qualitative Analysis}
Figure~\ref{fig:qualitative} shows two illustrative corrections made by rubric training. In the first case, the answer-only checkpoint counts the visible people correctly but then makes an unsupported age-to-birth-year inference; the rubric-trained checkpoint preserves the intermediate reasoning step. In the second, the answer-only checkpoint misreads the graph location, while the rubric-trained checkpoint traces the visible segments before answering. These examples illustrate the aggregate trend: rubric feedback rewards the intermediate visual and logical claims that scalar answer rewards often collapse into a single final verdict.

\paragraph{Where the gains come from.}
The gains are concentrated where benchmark success depends on preserving visual evidence through a reasoning chain. Rubric-based GRPO improves the knowledge average from $59.35$ to $61.88$ over answer-level GRPO, and the same pattern appears on MathVision, DynaMath, WeMath, LogicVista, and the visual-math/chart averages. These are not simply harder benchmarks; they are benchmarks where a correct final answer is often the end product of several fragile subclaims. A scalar reward can indicate whether a response is accepted, but it cannot distinguish an accurate visual interpretation followed by an invalid inference from an inaccurate visual interpretation followed by a correct answer reached by chance. V-Rubrics changes the unit of supervision from final-answer acceptability to grounded visual facts, valid reasoning steps, and satisfied instruction constraints. This better matches the error structure of multimodal reasoning: many failures are local, but their consequences appear only at the final answer.

\paragraph{Why rubrics improve credit assignment.}
Answer-level GRPO collapses distinct failures---a hallucinated chart value, an omitted constraint, or an inconsistent inference---into one scalar signal. Rubric scoring separates these cases into distinct item-level judgments, so a rollout can receive credit for grounded observations while still being penalized for the step that invalidates the answer. Global item-weight normalization keeps the rubric-reward scale comparable across examples, and prefix masking places item-level advantages on the response prefix that supports the verifier decision without renormalizing the active items at each token. This better aligns the credit signal with failure points in multimodal reasoning, especially on reasoning-heavy visual tasks. Additional discussion of prefix-localized credit and weaker benchmark regimes is given in Appendix~\ref{app:additional_analysis}.

\section{Conclusion}
\label{sec:conclusion}

We introduced Visual Rubrics-Based Reinforcement Learning, which treats visual faithfulness as a fine-grained credit-assignment problem rather than a post-hoc evaluation label. V-Rubrics 50K decomposes reference responses into visually grounded VF/RC/IF criteria and converts them into fine-grained scores and prefix-localized credit for VLM post-training. In Qwen3-VL-8B experiments with a shared SFT initialization and RL dataset, rubric-based GRPO improves over both the SFT baseline and scalar answer-level GRPO, especially on benchmarks that require grounded intermediate reasoning. More broadly, rubric-level supervision points toward visual grounding as a reusable training interface for multimodal alignment.

\section{Limitations}
\label{sec:limitations}

Our study has several limitations. First, V-Rubrics 50K depends on the quality of automatically generated rubrics and judge-model verification. Poorly specified criteria can encode reference-answer bias, visual ambiguity, or assumptions that are not fully supported by the image. Second, prefix-credit localization is approximate: the verifier sentence is aligned back to the response with a fuzzy match, so the resulting prefix credit should be interpreted as practical local feedback rather than exact token-level supervision. Finally, judge-model family bias may arise because Qwen-family judges score Qwen-family policies. Future work should evaluate rubric quality with larger human-audited sets, test judge diversity, and study transfer to other model families and safety-sensitive domains.

\section{Ethics}
\label{sec:ethics}

\paragraph{Data provenance.}
V-Rubrics 50K is constructed from publicly released visual question answering and visual reasoning datasets. The 17 source corpora listed in Appendix~\ref{app:data_construction} cover diagram, chart, document, mathematical, counting, educational, and general visual reasoning tasks. We do not crawl additional images. Each V-Rubrics 50K record contains the image payload used for training together with the instruction, reference answer, rubric annotations, difficulty metadata, and source metadata. The images originate from the 17 upstream datasets and remain subject to their respective licenses and usage terms; inclusion in V-Rubrics does not relicense them. Users must consult and comply with the upstream terms before using or redistributing the corresponding records.

\paragraph{Judge-model bias.}
Rubric items are produced by a large language model and scored by a separate judge model, so the reward signal can inherit biases from both the rubric generator and the verifier. The training-time rubric verifier receives the generated response and a self-contained criterion rather than the raw image; consequently, ambiguities or mistakes introduced when visual evidence is converted into criterion text can propagate directly into the reward. The verifier may also favor particular writing styles or reasoning templates. Self-contained, explicitly grounded criteria make these decisions auditable but do not eliminate judge bias. Downstream users should audit rubric distributions, verifier decisions, and failure cases before applying rubric-trained models beyond research settings.

\paragraph{Human annotation and deployment.}
V-Rubrics 50K is generated automatically and introduces no new demographic or sensitive-personal-attribute annotations. The trained policy is intended as a research artifact for studying visually grounded post-training, not as a standalone fact-checker or decision system.

\section{Acknowledgments}

This study is supported by the Ministry of Education, Singapore, under its MOE AcRF Tier 2 (MOE-T2EP20223-0002). This research is also supported by cash and in-kind funding from NTU S-Lab and industry partner(s).

\bibliography{custom}
\appendix

\section{Appendix Organization}
\label{sec:appendix}

Appendix~\ref{app:extended_context} surveys related work on visual reasoning, rubric judging, and hallucination-aware alignment. Appendix~\ref{app:data_construction} details the source coverage, rule-based filtering, difficulty composition, metadata, and annotation schema of V-Rubrics 50K. Appendix~\ref{app:training_details} specifies the SFT checkpoint provenance, GRPO configuration, reward construction, dense credit assignment, and decoding settings. Appendix~\ref{app:additional_analysis} analyzes the empirical effects of dense credit and the benchmark regimes in which it is less effective.

\section{Extended Context and Related Work}
\label{app:extended_context}

This section extends the related-work discussion to hallucination diagnostics and adjacent long-chain and video-reasoning systems, in addition to multimodal RL, rubric-based judging, and fine-grained alignment feedback.

\subsection{Expanded Related Work}
\label{app:more_related_work}

\paragraph{Open multimodal reasoning and RLVR.}
Open VLMs and reasoning-specialized post-training recipes provide the basis for the methods compared in Tables~\ref{tab:rubrics_main_general_knowledge} and~\ref{tab:rubrics_main_visual_math}. General-purpose backbones such as Qwen2.5-VL, Qwen3-VL, InternVL3, and LLaVA-OneVision provide strong perception, OCR, chart, and document understanding foundations~\citep{bai2025qwen25vl,bai2025qwen3vl,zhu2025internvl3}~\citep{li2024llavaonevision,an2025llavaonevision15}. Broader evaluations also test cross-disciplinary understanding and multi-hop multimodal agents~\citep{zou2026uni,tian2025mmina}. RLVR-style visual reasoning methods then adapt verifiable rewards to multimodal tasks, including Visual-RFT, VLM-R1, Vision-R1, Perception-R1, and Point-RFT~\citep{liu2025visualrft,shen2025vlmr1,huang2025visionr1}~\citep{xiao2025perceptionr1,ni2025pointrft}. Other reasoning-tuned systems add long-chain supervision, synthetic reasoning trajectories, or on-policy RL: VLAA-Thinker studies the tension between SFT imitation and subsequent RL; ThinkLite uses sample selection for data-efficient visual reasoning; VL-Rethinker incentivizes self-reflection; OpenVLThinker alternates SFT and RL; and MM-Eureka applies rule-based RL to multimodal STEM reasoning~\citep{chen2025vlaathinking,wang2025thinklite,wang2025vlrethinker}~\citep{deng2025openvlthinker,meng2025mmeureka}. Other recent systems emphasize scale, variance, or transfer: M2-Reasoning unifies general and spatial reasoning; MMR1 introduces variance-aware sampling; OVR transfers linguistic cognitive behaviors into visual reasoning; OpenMMReasoner provides an open general recipe and OMR checkpoints; and OneThinker extends image reasoning toward unified image/video reasoning~\citep{wang2025m2reasoning,leng2025mmr1,wei2025openvisionreasoner}~\citep{zhang2025openmmreasoner,feng2026onethinker}. These systems mostly optimize final-answer correctness or benchmark-level reward signals. V-Rubrics instead makes the reward object explicit by evaluating each response through grounded rubric items with interpretable dimensions and importance weights.

\paragraph{Long-chain visual reasoning and in-context adaptation.}
Insight-V and Insight-V++ are especially relevant because they treat visual reasoning as a long-chain process rather than a short answer-selection problem~\citep{dong2025insightv,dong2026insightvpp}. Insight-V generates structured long reasoning data and uses a reasoning/summary-agent design with preference optimization, while Insight-V++ extends the framework toward image-video reasoning and GRPO-style optimization. Demo-ICL and Ego-R1 study complementary adaptation settings: Demo-ICL evaluates whether multimodal models can learn procedural video knowledge from demonstrations in context, while Ego-R1 trains a tool-using RL agent for ultra-long egocentric video reasoning~\citep{dong2026demoicl,tian2025egor1}. Adjacent work extends structured multimodal reasoning to test-time scaling and video generation~\citep{liangyu2026unit,huang2026vchain}. These works broaden the design space around reasoning trajectories, adaptation, and evaluator-guided improvement. Our work intersects with them in its use of structured supervision, but the supervision target differs: we decompose reference answers into visually checkable atomic propositions and use those propositions as dense credit during RL.

\paragraph{Rubrics, judges, and reward models.}
Rubric-style evaluation grew out of the observation that a single scalar score often hides which part of a response succeeded or failed. Text evaluation work such as G-Eval, Prometheus 2, and LLM-Rubric uses structured criteria to improve judge reliability~\citep{liu-etal-2023-g,kim-etal-2024-prometheus2,hashemi-etal-2024-llm}. Multimodal evaluation and reward modeling extends this idea through per-sample criteria, VLM judges, preference critics, and learned reward models~\citep{ge2023mllmbench,lee-etal-2024-prometheus-vision,xiong2025llavacritic}~\citep{wang2025skyworkvlreward}. Adjacent visual-generation work explores promptable evaluation agents~\citep{zhang2025evaluation,tian2026open}. Recent reward-model benchmarks and training recipes further show that multimodal reward quality is itself a difficult evaluation target~\citep{li2024vlrewardbench,yasunaga2025multimodalrewardbench,zhang2025r1reward}. Atomic factuality work such as FActScore and TIFA further motivates decomposing outputs into checkable units~\citep{min-etal-2023-factscore,hu2023tifa}. V-Rubrics follows this criterion-based line, but turns the criteria into training-time item credit rather than only using them as post-hoc judging text.

\paragraph{Visual faithfulness and hallucination-aware alignment.}
Hallucination-aware methods attack unsupported visual claims from the evaluation and alignment sides. CHAIR, POPE, HallusionBench, AMBER, and FaithScore diagnose object hallucinations, visual illusions, and atomic image-fact errors~\citep{rohrbach-etal-2018-object,li-etal-2023-evaluating,guan2024hallusionbench}~\citep{wang2023amber,jing-etal-2024-faithscore}. Alignment methods such as LLaVA-RLHF, RLHF-V, fine-grained AI feedback, HDPO, and RLAIF-V show that hallucination-sensitive feedback can improve multimodal reliability~\citep{sun-etal-2024-aligning,yu2024rlhfv,xiao2024fgaif}~\citep{fu-etal-2025-hdpo,yu2025rlaifv}. V-Rubrics combines these threads by converting visually grounded rubric judgments into an RL reward and pairing them with example-level \texttt{rs\_score}-derived difficulty metadata rather than treating them only as evaluation artifacts.

\paragraph{Difficulty-aware sample composition.}
Prior work on sample selection and self-paced organization motivates tracking example difficulty when assembling training data~\citep{kumar2010selfpaced}. V-Rubrics 50K stores an example-level \texttt{rs\_score}, derives the corresponding difficulty category deterministically, and uses a fixed mixture of 18,121 hard, 25,306 medium, and 6,821 simple examples. This connects the composition metadata in Appendix~\ref{app:data_construction} with the training details in Appendix~\ref{app:detailed_reward_training}.

\section{V-Rubrics 50K Data and Rubric Construction}
\label{app:data_construction}

This section describes the source selection, rule-based filtering, difficulty composition, and annotation schema used to construct V-Rubrics 50K.

\subsection{Source Datasets and Final Inventory}
\label{app:data_sources}

The final V-Rubrics release contains 50,248 examples drawn from 17 canonical training sources: AI2D~\citep{kembhavi2016ai2d}, ChartGalaxy~\citep{li2025chartgalaxy}, ChartNet~\citep{kondic2026chartnet}, ChartQA~\citep{masry2022chartqa}, ChartQA-X~\citep{hegde2025chartqax}, ChartX~\citep{xia2024chartx}, DocVQA~\citep{mathew2021docvqa}, Geometry3K from Inter-GPS~\mbox{\citep{lu2021intergps}}, GeoSynth Choice~\citep{pan2025geogen}, InfographicVQA~\citep{mathew2021infographicvqa}, MM-K12~\citep{du2025mmprm}, PixMo-Count~\citep{deitke2025molmo}, ThinkLite and ThinkLite-Hard~\citep{wang2025thinklite}, ViRL39K~\citep{wang2025vlrethinker}, and We-Math2.0-STD and We-Math2.0-Pro~\citep{qiao2025wemath2}. We report canonical sources rather than collapsing project families, so the two We-Math 2.0 sources are listed separately, following the same convention used for ThinkLite and ThinkLite-Hard. Together, the 17 canonical sources cover complementary visual skills, including diagram understanding, chart and document question answering, mathematical visual reasoning, counting, and educational multimodal reasoning.

\subsection{Data Construction and Difficulty Composition}
\label{app:data_filtering}

We construct V-Rubrics 50K through a single forward pipeline. Starting from the 17 canonical training sources, Stage~1 applies rule-based checks before rejection sampling: records must have valid required fields and media, present a non-trivial learning target, meet basic language-quality requirements, and pass identity and strict-content deduplication. Stage~2 uses rejection sampling to support example selection and derive difficulty from the resulting example-level scores. We then form the source and difficulty composition reported below, generate rubric annotations, and use the resulting 50,248 records for RL training.

The resulting inventory contains 50,248 unique UIDs and 50,248 unique strict question--answer--image content fingerprints. These identity, content, and media checks are construction invariants applied before a record enters V-Rubrics 50K.

Difficulty is represented by an example-level hard, medium, or simple assignment derived from the stored rejection-sampling score, \texttt{rs\_score}$=k/8$: $0/8$ maps to hard, $1/8$ through $5/8$ map to medium, and $6/8$ through $7/8$ map to simple. We discarded samples with a score of $8/8$; accordingly, the release contains no $8/8$ examples. The fixed dataset composition contains 18,121 hard (36.1\%), 25,306 medium (50.4\%), and 6,821 simple examples (13.6\%). These derived assignments describe the difficulty-stratified mixture used for training; difficulty is not a field of individual rubric items. Percentages are rounded independently to one decimal place.

\begin{table*}[t]
\centering
\small
\begin{tabular}{lrr}
\toprule
Source & Final examples & Share \\
\midrule
AI2D & 1,758 & 3.5\% \\
ChartGalaxy & 1,289 & 2.6\% \\
ChartNet & 681 & 1.4\% \\
ChartQA & 687 & 1.4\% \\
ChartQA-X & 1,314 & 2.6\% \\
ChartX & 1,203 & 2.4\% \\
DocVQA & 336 & 0.7\% \\
Geometry3K & 919 & 1.8\% \\
GeoSynth Choice & 1,290 & 2.6\% \\
InfographicVQA & 2,071 & 4.1\% \\
MM-K12 & 6,153 & 12.2\% \\
PixMo-Count & 2,121 & 4.2\% \\
ThinkLite-Hard & 8,624 & 17.2\% \\
ThinkLite & 13,041 & 26.0\% \\
ViRL39K & 6,188 & 12.3\% \\
We-Math2.0-STD & 1,076 & 2.1\% \\
We-Math2.0-Pro & 1,497 & 3.0\% \\
\midrule
Total & 50,248 & 100.0\% \\
\bottomrule
\end{tabular}
\caption{\textbf{Canonical source inventory of V-Rubrics 50K.} Counts are computed from the 50,248 records in the current release; source shares are rounded independently to one decimal place.}
\label{tab:vrubrics_source_counts}
\end{table*}

\subsection{Rubric Metadata and Generation Schema}
\label{app:rubric_metadata}
\label{app:rubric_prompt}

Each example is associated with its source image, instruction, reference answer, canonical source, and example-level difficulty assignment. Rubric construction decomposes each reference answer into short, independently checkable atomic propositions. Each proposition is tagged as VF, RC, or IF and receives an importance label and numeric weight used by the reward function. The rubric schema contains no item-level difficulty field. Across V-Rubrics 50K, 352,938 rubric items comprise 209,436 VF items (59.3\%), 101,369 RC items (28.7\%), and 42,133 IF items (11.9\%); percentages are rounded independently to one decimal place.

All rubric annotations are generated by \texttt{Gemini-3-Pro}~\citep{google2026gemini3pro} under one structured, image-conditioned protocol. For every record, the model receives the source image, instruction, and reference answer directly and returns a list of independently checkable evaluation criteria using the four-field schema below. The prompt is designed to make each annotation usable as a reward signal rather than only as a post-hoc evaluation note. It therefore asks for atomic propositions, explicit visual evidence when an item depends on the image, categorical importance labels, signed weights, and a dimension label. This design supports the four principles in Section~\ref{sec:rubric_design}: visual grounding is enforced by requiring concrete image facts; self-containment is enforced by requiring the criterion to be judgeable without external context; coverage is encouraged by asking for multiple criteria rather than a single holistic judgment; and importance is represented by the \textsc{Essential}, \textsc{Important}, \textsc{Optional}, and \textsc{Pitfall} prefixes.

The annotation schema uses three capability dimensions: \textbf{Visual Faithfulness (VF)}---whether the response accurately reflects what is visually present in the image; \textbf{Reasoning Consistency (RC)}---whether the response draws logically valid conclusions and inferences from the observed visual information; and \textbf{Instruction Following (IF)}---whether the response adheres to the format, style, and task requirements specified in the prompt. For VQA tasks, the generation prompt specifies the priority $\mathrm{VF}\!\gg\!\mathrm{RC}\!>\!\mathrm{IF}$ so that visual grounding remains the dominant evaluation axis. Table~\ref{tab:rubric_generation_prompt} presents the core generation instructions and four-field output schema.

\begin{table*}[t]
\centering
\begin{minipage}[t]{0.98\textwidth}
\begin{lstlisting}[basicstyle=\ttfamily\tiny,frame=single,xleftmargin=0pt,xrightmargin=0pt]
You are an expert rubric writer. Your job is to generate a self-contained set of evaluation criteria
("rubrics") for judging how good a response is to a given question with respect to an image. Rubrics
can cover aspects of a response such as, but not limited to, factual correctness, visual faithfulness,
visual grounding, reasoning faithfulness, reasoning depth, clarity, completeness, ideal-response
characteristics, style, helpfulness, harmlessness, contextual relevance.

Each rubric item must be an atomic proposition, where the statement can be evaluated as strictly True
or False without external context by non-expert readers, and if the response needs the information
from the image, the rubric item should clearly state the content / info included in the image, like
the numbers, texts, etc.

Each rubric item should fall into one of 3 capability types:
- Visual Faithfulness (VF): whether the response accurately reflects what is visually present in the
  image.
- Reasoning Consistency (RC): whether the response draws logically valid conclusions and inferences
  from the observed visual information.
- Instruction Following (IF): whether the response adheres to the format, style, and task requirements
  specified in the prompt.
In VQA tasks, usually VF >> RC > IF. Ensure that VF rubric items constitute the majority, since visual
grounding is the primary competency being evaluated.

Input:
- question: the question to be answered.
- image: the image to be used for answering the question.
- reference_answer: an ideal but not necessarily exhaustive reference; treat as guidance only.

Output:
A JSON list of rubric objects, where each object has exactly four keys:

{
  "name": "a concise, descriptive label that acts as a unique identifier, e.g., Visual_Accuracy",
  "description": "a short, clear criterion beginning with the category prefix Essential / Important /
                  Optional / Pitfall",
  "weight": "1-5 for Essential / Important / Optional (5 = most important); -1 or -2 for Pitfall",
  "type": "VF (= Visual Faithfulness) / RC (= Reasoning Consistency) / IF (= Instruction Following)"
}

Category prefixes (meaning):
- Essential: the validity floor of the response; once violated, the response is functionally useless.
- Important: defines reasoning or answer quality; once missed, the response may be partially correct
  but poor.
- Optional: useful but non-critical, e.g., stylistic preference or formatting nicety.
- Pitfall: a common mistake or hallucination that should be avoided.

Formatting rules:
- When referring to answers, be explicit: use "Identifies(A)" for multiple-choice, or state the
  expected content ("States the answer is 42") for free-form. Never use vague phrasing like "gives
  the correct answer".
- Do not copy large blocks of the question or reference_answer into rubric descriptions. Each
  description must begin with its category prefix; no extra keys are allowed.
- Avoid vague language: instead of "clearly states what the chart shows", explicitly state the
  critical visual content, e.g., "the chart shows that revenue increased by 80% in February".
- For VQA tasks, prioritize VF rubrics so they form the majority.

Now, given the question and reference_answer, generate the rubric as described. The reference answer
is an ideal response but not necessarily exhaustive; use it only as guidance. Do not output anything
else.
\end{lstlisting}
\end{minipage}
\caption{\textbf{Core rubric-generation instructions and output schema used to construct V-Rubrics 50K.}}
\label{tab:rubric_generation_prompt}
\end{table*}

\section{Reward, Training, and Decoding Details}
\label{app:training_details}

This section documents the SFT checkpoint, the complete GRPO objective, component-wise prefix-localized credit assignment, and evaluation-time generation settings.

\subsection{SFT Checkpoint Provenance}
\label{app:sft_provenance}

Our training pipeline begins with Qwen3-VL-8B-Instruct and adapts the cold-start data recipe of OpenMMReasoner~\citep{zhang2025openmmreasoner}. Specifically, we train on the OpenMMReasoner-SFT-874K dataset, whose five configurations are \texttt{llava\_cot}, \texttt{m1\_sft}, \texttt{mmr1}, \texttt{OpenVLThinker-sft-iter3}, and \texttt{WeMath}. The ``SFT'' entries in the result tables refer to the resulting Qwen3-VL-8B checkpoint, denoted $\pi_{\mathrm{SFT}}$, rather than the released OMR-7B-ColdStart checkpoint. For each GRPO variant, the actor $\pi_{\theta^{(0)}}$ and the frozen KL reference policy $\pi_{\mathrm{ref}}$ are independently initialized from $\pi_{\mathrm{SFT}}$:
\[
\pi_{\theta^{(0)}}=\pi_{\mathrm{ref}}=\pi_{\mathrm{SFT}},
\qquad \pi_{\mathrm{ref}}\ \text{frozen}.
\]
Starting from this common initialization, the answer-level and rubric-based runs optimize separate actor copies on $\mathcal{D}_{\text{V-Rubrics}}$ while keeping their respective reference copies fixed.

\subsection{GRPO Training Configuration}
\label{app:hyperparams}

Table~\ref{tab:hyperparams} reports the GRPO configuration of the runs in Tables~\ref{tab:rubrics_main_general_knowledge} and~\ref{tab:rubrics_main_visual_math}. Both variants use the same SFT initialization, the same V-Rubrics 50K examples, the same rollout group size, and the same maximum training horizon; they differ in the train and PPO mini-batch sizes shown separately in the table, and in the reward and credit-assignment mechanism of Section~\ref{sec:method}. All other values are shared.

\begin{table*}[t]
\centering
\small
\begin{tabular}{@{}p{0.30\textwidth}p{0.64\textwidth}@{}}
\toprule
\textbf{Setting} & \textbf{Configuration} \\
\midrule
\multicolumn{2}{@{}l}{\textit{Initialization and data}} \\
\addlinespace[2pt]
Backbone before SFT & Qwen3-VL-8B-Instruct \\
SFT corpus & OpenMMReasoner-SFT-874K \\
RL corpus & V-Rubrics 50K \\
Actor initialization & $\pi_{\theta^{(0)}}=\pi_{\mathrm{SFT}}$; $\pi_\theta$ is trainable \\
Reference policy & $\pi_{\mathrm{ref}}=\pi_{\mathrm{SFT}}$ at initialization and frozen thereafter \\
\midrule
\multicolumn{2}{@{}l}{\textit{Optimization and regularization}} \\
\addlinespace[2pt]
Trainer & verl GRPO (Megatron backend) \\
Advantage estimator & group-relative (GRPO) \\
Optimizer & Megatron Adam; weight decay $0.01$ \\
Learning rate & $2\!\times\!10^{-6}$ \\
Train batch size & $192$ (rubric-based); $480$ (answer-level) \\
PPO mini-batch size & $192$ (rubric-based); $96$ (answer-level) \\
KL regularization & low-variance KL in the actor loss; coefficient $0.01$ \\
Entropy coefficient & $0$ \\
Semantic answer / rubric balance & $0.5$ / $0.5$ (rubric-based run only) \\
Seed & $42$ \\
\midrule
\multicolumn{2}{@{}l}{\textit{Rollout and systems settings}} \\
\addlinespace[2pt]
Max prompt length & $8{,}192$ tokens \\
Max response length & $8{,}192$ tokens \\
Rollouts per prompt $G$ & $12$ \\
Maximum epochs & $5$ \\
Tensor / pipeline / context parallel & $1$ / $1$ / $2$ \\
GPUs per node & $8$ \\
Checkpoint / evaluation interval & every $20$ / $20$ steps \\
Rubric evaluation & independent per-item judgments, evaluated concurrently \\
Rubric judge & Qwen3-VL-235B-A22B (vLLM, FP8) \\
Answer Equivalence judge & Qwen3-VL-235B-A22B (vLLM, FP8) \\
\bottomrule
\end{tabular}
\caption{\textbf{GRPO training configuration.} Settings are shared by the rubric-based run and the answer-level baseline except where a row lists both variants separately.}
\label{tab:hyperparams}
\end{table*}

\subsection{GRPO Objective and Policy Update}
\label{app:grpo_optimization}

Let $\mathcal{D}=\mathcal{D}_{\text{V-Rubrics}}$ denote the RL training set. The scalar sequence-level variant optimizes the expected blended reward
\begin{equation}
\max_\theta\;
\mathbb{E}_{x\sim\mathcal{D},\,a\sim\pi_\theta(\cdot\mid x)}
\big[\,R(a,x)\,\big].
\label{eq:rl_objective}
\end{equation}
For each $x$, the rollout policy samples $G$ tokenized responses
\begin{equation}
\begin{aligned}
a^{(g)}&=(a_1^{(g)},\ldots,a_{T_g}^{(g)}),\\
a^{(g)}&\sim\pi_{\theta_{\mathrm{old}}}(\cdot\mid x),
\qquad g\in\{1,\ldots,G\},
\end{aligned}
\end{equation}
where $T_g$ is the valid response length. Given a token-level advantage $A_t^{(g)}$, the policy maximizes the standard clipped surrogate
\begin{equation}
\begin{aligned}
\bar{\varrho}_t^{(g)}
&=\mathrm{clip}\!\left(
\varrho_t^{(g)},1\!-\!\epsilon_c,1\!+\!\epsilon_c
\right),\\
\mathcal{J}_{\mathrm{clip}}(\theta)
&=\mathbb{E}_{x,g,t}\!\left[
\min\!\left(
\varrho_t^{(g)}A_t^{(g)},
\bar{\varrho}_t^{(g)}A_t^{(g)}
\right)
\right],
\end{aligned}
\label{eq:grpo_surrogate}
\end{equation}
where
\begin{equation}
\varrho_t^{(g)}
=
\frac{\pi_\theta(a_t^{(g)}\mid x,a_{<t}^{(g)})}
{\pi_{\theta_{\mathrm{old}}}(a_t^{(g)}\mid x,a_{<t}^{(g)})}
\end{equation}
is the token importance ratio and $\epsilon_c$ is the clipping radius. Here $\pi_{\theta_{\mathrm{old}}}$ is the pre-update actor snapshot, whereas $\pi_{\mathrm{ref}}$ is the frozen KL reference. Training uses a low-variance KL term in the actor loss, no entropy bonus, and \texttt{verl}'s dual-clipped treatment of negative advantages. Missing judgments and retry behavior are specified in Appendix~\ref{app:decoding}.

\subsection{Reward and Advantage Composition}
\label{app:advantage_composition}

In the component-wise, prefix-localized variant (\S\ref{sec:dense_credit}), the answer parser uses the final \texttt{<answer>...</answer>} payload when present and otherwise passes the full response to the equivalence judge. Its standardized advantage is weighted by $\alpha=0.5$ and broadcast to all valid tokens; each positive rubric contributes a separately standardized advantage within the remaining budget $1-\alpha$ (Equation~\ref{eq:dense_credit}).

Equation~\ref{eq:group_baseline} uses the sample standard deviation for the answer and the population standard deviation for each rubric's scored subset; a constant component provides no group-relative signal. Item weights are normalized once, with no token-dependent renormalization after the prefix mask in Equation~\ref{eq:prefix_mask}. The scalar variant instead aggregates $R_{\mathrm{rub}}$ with $R_{\mathrm{ans}}$ before group standardization.

\subsection{Detailed Reward and Training Formulation}
\label{app:detailed_reward_training}

\paragraph{Rubric labels, aligned scores, and dimensions.}
Positive-weight \textsc{Essential}, \textsc{Important}, and \textsc{Optional} criteria grant partial credit; negative-weight \textsc{Pitfall} criteria describe failures to avoid. The aligned score $s_j\in\{0,1\}$ equals 1 when a criterion is satisfied, including when a pitfall is avoided. Positive weights set item importance; a confirmed \textsc{Pitfall} violation ($s_j=0$) vetoes answer and positive-rubric credit, while transport or parsing failures do not. \textbf{Visual Faithfulness (VF)} checks image evidence, \textbf{Reasoning Consistency (RC)} checks conclusions from that evidence, and \textbf{Instruction Following (IF)} checks task and format requirements.

\paragraph{Difficulty composition.}
Rubric-based GRPO uses the fixed mixture of 18,121 hard, 25,306 medium, and 6,821 simple examples. Rubric items contain only \texttt{name}, \texttt{description}, \texttt{weight}, and \texttt{type}; difficulty derives from the example-level \texttt{rs\_score}, not from rubric items or a training schedule.

\paragraph{Prefix-credit localization.}
For each positive rubric item $r_j$, the verifier returns $s_j$ and, when available, a verbatim scoring sentence. We align that sentence to the tokenized response with a partial-ratio fuzzy match and denote the matched endpoint by $t_{j,\mathrm{end}}^{(g)}$. The item advantage applies to tokens $t\le t_{j,\mathrm{end}}^{(g)}$; an unmatched scored item uses $T_g$ and therefore spans the response.

This mask is a coarse heuristic: earlier tokens may receive credit from several later items. We retain globally normalized weights without active-item renormalization, while answer equivalence remains sequence-level. Rubric credit can therefore vary within a response while rollout comparison remains standard GRPO.

\paragraph{Information granularity.}
Rubric aggregation preserves partial correctness; component-wise standardization retains item-level differences, and prefix masks determine where they apply.

\subsection{Decoding Settings}
\label{app:decoding}

\paragraph{RL rollouts.} vLLM generates $G=12$ samples per prompt with temperature $1.0$, top-$p$ disabled, and a maximum response length of $8{,}192$ tokens. Judges use temperature $0.0$ and output caps of $8$ tokens for answer equivalence and scalar rubric judgments, or $8{,}192$ for prefix-credit judgments. Failed calls use up to $3$ exponential-backoff retries. An unresolved answer score becomes zero; a failed positive-item judgment becomes zero in the scalar variant and is omitted from prefix component statistics. Weights are not rescaled, and a failed \textsc{Pitfall} judgment is not a confirmed violation.

\paragraph{Evaluation.} Locally evaluated checkpoints in Tables~\ref{tab:rubrics_main_general_knowledge} and~\ref{tab:rubrics_main_visual_math} use VLMEvalKit and vLLM. Qwen3-VL chain-of-thought generation uses \texttt{THINKING=True} and \texttt{SPLIT\_THINK=True}, with benchmark-specific final-answer parsers. We use VLMEvalKit's default temperatures and a separate vLLM answer-judge endpoint.

\section{Additional Analysis}
\label{app:additional_analysis}

\paragraph{What dense credit adds.}
Table~\ref{tab:ablations} rises from $66.25$ with answer-only GRPO to $67.74$ with scalar rubric aggregation and $68.04$ with component-wise prefix credit. The $0.30$-point gap between rubric variants reflects component-wise standardization and localization jointly, not localization alone. Factorized prefix credit can reduce irrelevant credit or blame across mixed observations, reasoning, formatting, and later mistakes.

\paragraph{Where the method is weaker.}
Answer-level GRPO remains stronger on MMBench-Dev, MathVerse V/O, and CharXiv reasoning, so rubric shaping is not uniformly aligned with every metric. This likely reflects rubric-metric mismatch: some outcomes depend on omitted or misweighted requirements such as exact symbolic correctness, option normalization, or dataset-specific conventions. V-Rubrics is strongest when benchmark success depends on grounded subclaims that its rubrics explicitly score.

\end{document}